\documentclass[10pt,twocolumn,letterpaper]{article}

\usepackage{cvpr}              

\usepackage{graphicx}
\usepackage{amsmath}
\usepackage{amssymb}
\usepackage{booktabs}
\usepackage{CJKutf8}

\usepackage[pagebackref,breaklinks,colorlinks]{hyperref}

\usepackage[capitalize]{cleveref}
\crefname{section}{Sec.}{Secs.}
\Crefname{section}{Section}{Sections}
\Crefname{table}{Table}{Tables}
\crefname{table}{Tab.}{Tabs.}

\def\cvprPaperID{4387} 
\def\confName{CVPR}
\def\confYear{2023}

\begin{document}

\title{DeepVecFont-v2:\\ Exploiting Transformers to Synthesize Vector Fonts with Higher Quality}

\author{Yuqing Wang$^{1,2}$, Yizhi Wang$^1$, Longhui Yu$^2$, Yuesheng Zhu$^2$, Zhouhui Lian$^1$\thanks{Corresponding author. E-mail: lianzhouhui@pku.edu.cn}\\
$^1$Wangxuan Institute of Computer Technology, Peking University, China\\
$^2$School of Electronic and Computer Engineering, Peking University, China\\
}

\maketitle

\begin{abstract}
    
    Vector font synthesis is a challenging and ongoing problem in the fields of Computer Vision and Computer Graphics.
    The recently-proposed DeepVecFont~\cite{wang2021deepvecfont} achieved state-of-the-art performance by exploiting information of both the image and sequence modalities of vector fonts. However, it has limited capability for handling long sequence data and heavily relies on an image-guided outline refinement post-processing. Thus, vector glyphs synthesized by DeepVecFont still often contain some distortions and artifacts and cannot rival human-designed results. To address the above problems, this paper proposes an enhanced version of DeepVecFont mainly by making the following three novel technical contributions. First, we adopt Transformers instead of RNNs to process sequential data and design a relaxation representation for vector outlines, markedly improving the model's capability and stability of synthesizing long and complex outlines. Second, we propose to sample auxiliary points in addition to control points to precisely align the generated and target Bézier curves or lines. Finally, to alleviate error accumulation in the sequential generation process, we develop a context-based self-refinement module based on another Transformer-based decoder to remove artifacts in the initially synthesized glyphs. Both qualitative and quantitative results demonstrate that the proposed method effectively resolves those intrinsic problems of the original DeepVecFont and outperforms existing approaches in generating English and Chinese vector fonts with complicated structures and diverse styles.

\end{abstract}


\section{Introduction}


Vector fonts, in the format of Scalable Vector Graphics (SVGs), are widely used in displaying documents, arts, and media contents. However, designing high-quality vector fonts is time-consuming and costly, requiring extensive experience and professional skills from designers.
Automatic font generation aims to simplify and facilitate the font designing process: learning font styles from a small set of user-provided glyphs and then generating the complete font library. until now, there still exist enormous challenges due to the variety of topology structures, sequential lengths, and styles, especially for some writing systems such as Chinese.

\begin{figure}[t!]
    \includegraphics[width=\columnwidth]{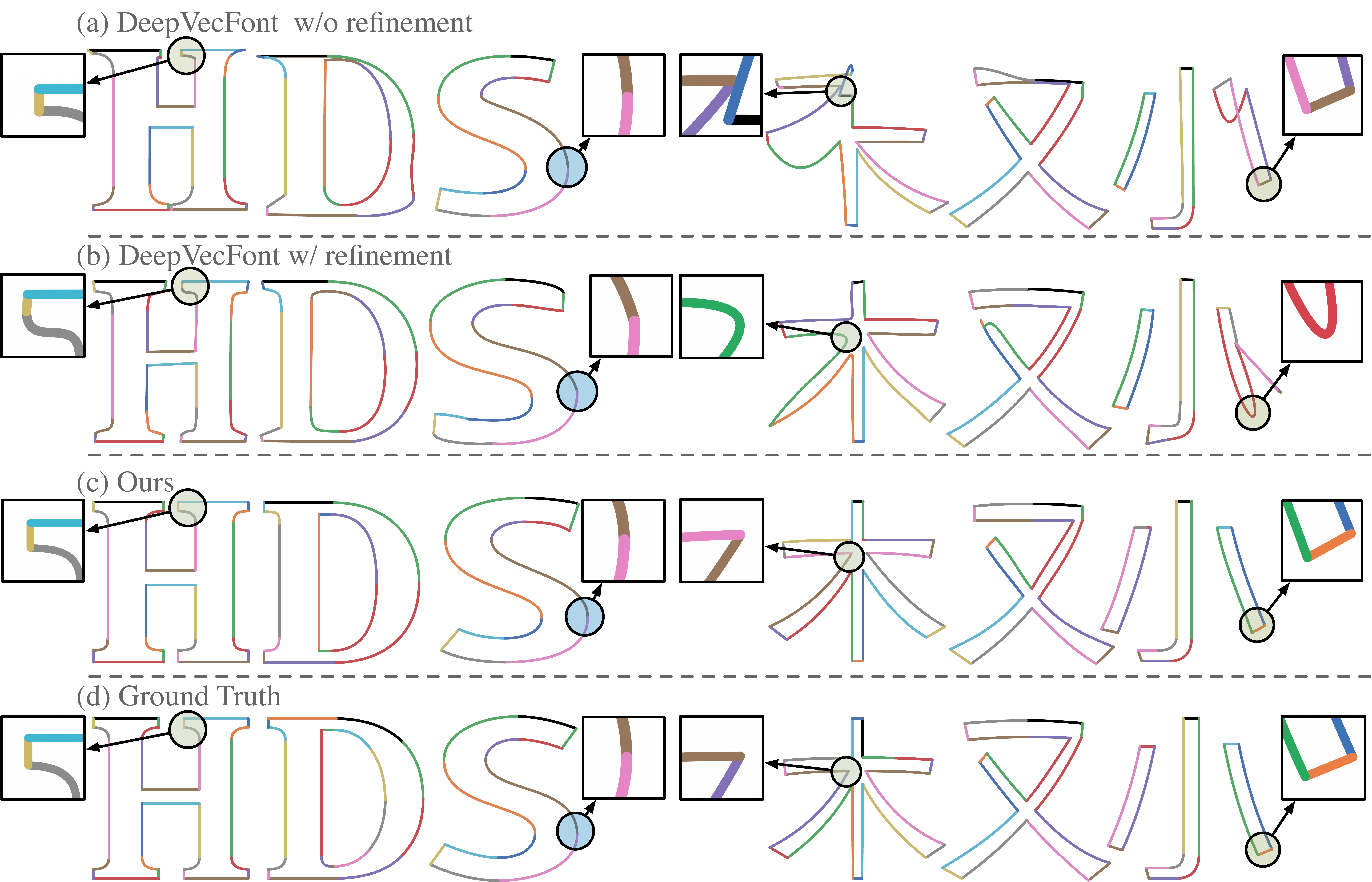}
       \caption{Visualization of the vector glyphs synthesized by DeepVecFont and Ours, where different colors denote different drawing commands. (a) DeepVecFont w/o refinement suffers from location shift. (b) DeepVecFont w/ refinement has both over-smoothness (see green circles) and under-smoothness (see blue circles). (c) Our method can directly synthesize visually-pleasing results with compact and coordinated outlines. Zoom in for better inspection.}
    
    \label{page1}
    \end{figure}

Recent years have witnessed significant progress~\cite{svgvae, carlier2020deepsvg} made by deep learning-based methods for vector font generation. Nevertheless, vector fonts synthesized by these existing approaches often contain severe distortions and are typically far from satisfactory. More recently, Wang and Lian~\cite{wang2021deepvecfont} proposed DeepVecFont that utilizes a dual-modality learning architecture by exploiting the features of both raster images and vector outlines to synthesize visually-pleasing vector glyphs and achieve state-of-the-art performance.
However, DeepVecFont tends to bring location shift to the raw vector outputs (Fig. \ref{page1}(a)), which are then further refined according to the synthesized images. Specifically, DeepVecFont adopts a differentiable rasterizer~\cite{2020diffvg} to fine-tune the coordinates of the raw vector glyphs by aligning their rasterized results and the synthesized images. However, after the above process, the refined vector outputs tend to be over-fitted to the inherent noise in the synthesized images. Thus, there often exist suboptimal outlines (Fig. \ref{page1}(b)) with over-smoothed corners (green circles) or under-smoothed adjacent connections (blue circles) in the final synthesized vector fonts, making them unsuited to be directly used in real applications.

To address the above-mentioned issues, we propose a new Transformer-based~\cite{2017Attentionisallyouneed} encoder-decoder architecture, named DeepVecFont-v2, to generate high-fidelity vector glyphs with compact and coordinated outlines. 
Firstly, we observed that the commonly used SVG representation, which shares the same starting and ending points between adjacent drawing commands, is more suitable for the learning process of RNNs~\cite{1997Lstm} than Transformers~\cite{2017Attentionisallyouneed}.
RNNs simulate a recurrent drawing process, where the next movement is determined according to the current hidden state fed with the current drawing command. Therefore, the starting point of the following drawing command can be omitted (replaced by the ending point of the current drawing command).
On the contrary, Transformers make drawing prediction based on the self-attention operations performed on any two drawing commands, whether adjacent or not. Therefore, to make the attention operator receive the complete information of their positions, the starting point of each drawing command cannot be replaced by the ending point of the previous command. Based on the above observation, we propose a relaxation representation that models these two points separately and merges them via an extra constraint.

Secondly, although the control points of a Bézier curve contain all the primitives, we found that the neural networks still need more sampling points from the curve to perform a better data alignment.
Therefore, we sample auxiliary points distributed along the Bézier curves when computing the proposed Bézier curve alignment loss. 

Thirdly, to alleviate the error accumulation in the sequential generation process, we design a self-refinement module that utilizes the context information to further remove artifacts in  the initially synthesized results.

Experiments conducted on both English and Chinese font datasets demonstrate the superiority of our method in generating complicated and diverse vector fonts and its capacity for synthesizing longer sequences compared to existing approaches. To summarize, the major contributions of this paper are as follows:

\begin{itemize} 
    \item[-] We develop a Transformer-based generative model, accompanied by a relaxation representation of vector outlines, to synthesize high-quality vector fonts with compact and coordinated outlines.
    \item[-] We propose to sample auxiliary points in addition to control points to precisely align the generated and target
outlines, and design a context-based self-refinement module to
fully utilize the context information to further remove artifacts.
    \item[-] Extensive experiments have been conducted to verify that state-of-the-art performance can be achieved by our method in both English and Chinese vector font generation.
\end{itemize}

\section{Related Work}

Font generation methods can be roughly classified into approaches that aim to synthesize fonts consisting of raster images and vector glyphs, respectively.

In recent years, glyph image synthesis methods typically draw inspirations from recent advances in deep generative models such as VAEs~\cite{2014Autoencoder} and GANs~\cite{2014Generative}.
Tian et al.~\cite{tian2017zi2zi} and Lyu et al.~\cite{lyu2017auto} employed the framework of pix2pix~\cite{pix2pix} based on cGAN~\cite{mirza2014conditional} to transfer a template font to target fonts with desired styles. Zhang et al.~\cite{emd} proposed EMD to explicitly separate the style and content features of glyph images.
Mc-GAN~\cite{azadi2018multi} and AGIS-Net~\cite{gao2019agisnet} decomposed the pipeline of artistic font generation into glyph image synthesis and texture transfer.
Wang et al.~\cite{wang2020attribute2font} presented Attribute2font, a cGAN-based network to synthesize fonts according to user-specified attributes and their corresponding values. 
For Chinese font generation, Park et al.~\cite{park2021few} and Kong et al.~\cite{kong2022look} proposed to fully exploit the component and layout information in Chinese glyphs.
Xie et al.~\cite{xie2021dg} proposed a deformable generative network which can synthesize target glyph images without direct supervision.
Tang et al.~\cite{tang2022fewshot} adopted a cross-attention mechanism for a fine-grained local style representation to generate target glyph images. \par
There is growing interest in the task of vector font synthesis, which can directly deliver results with scale-invariant representation.
Suveeranont and Igarash~\cite{suveeranont2010example} proposed to represent the user-defined character example as a weighted sum of the outlines and skeletons from the template fonts, and apply the weights to all characters to generate the new font. Campbell and Kautz~\cite{fontmf} aimed to learn a font manifold from existing fonts and create new font styles by interpolation and extrapolation from the manifold. SketchRNN~\cite{ha2017sketchrnn} employed a sequence VAE based on bi-directional RNNs 
to generate sketches. Easyfont~\cite{lian2018easyfont} decomposed the handwriting style into the shape and layout style of strokes to generate personal handwriting Chinese characters. SVG-VAE~\cite{svgvae} developed an image autoencoder architecture to learn style vectors of fonts, and then used LSTMs~\cite{1997Lstm} followed by a Mixture Density Network~\cite{1994Mixture} to generate the SVG drawing sequence. DeepSVG~\cite{carlier2020deepsvg} adopted a hierarchical generative network based on Transformers to generate vector icons with multiple paths. Im2Vec~\cite{reddy2021Im2Vec} can directly generate vector graphics from raster training images without explicit vector supervision. DeepVecFont~\cite{wang2021deepvecfont} exploited both the image and sequence modalities to facilitate the synthesis of vector glyphs. Liu et al.~\cite{implicit} proposed to represent glyphs implicitly as shape primitives enclosed by several quadratic curves, followed by a style transfer network to render glyph images at arbitrary resolutions. More recently, Aoki and Aizawa~\cite{svgforcn} introduced AdaIn \cite{adain} into the pipeline of DeepSVG to synthesize Chinese vector fonts. However, there still often exist non-negligible distortions and artifacts on the vector glyphs synthesized by the above-mentioned existing approaches. 

\begin{figure*}[t!]
\begin{center}
\includegraphics[width=\textwidth]{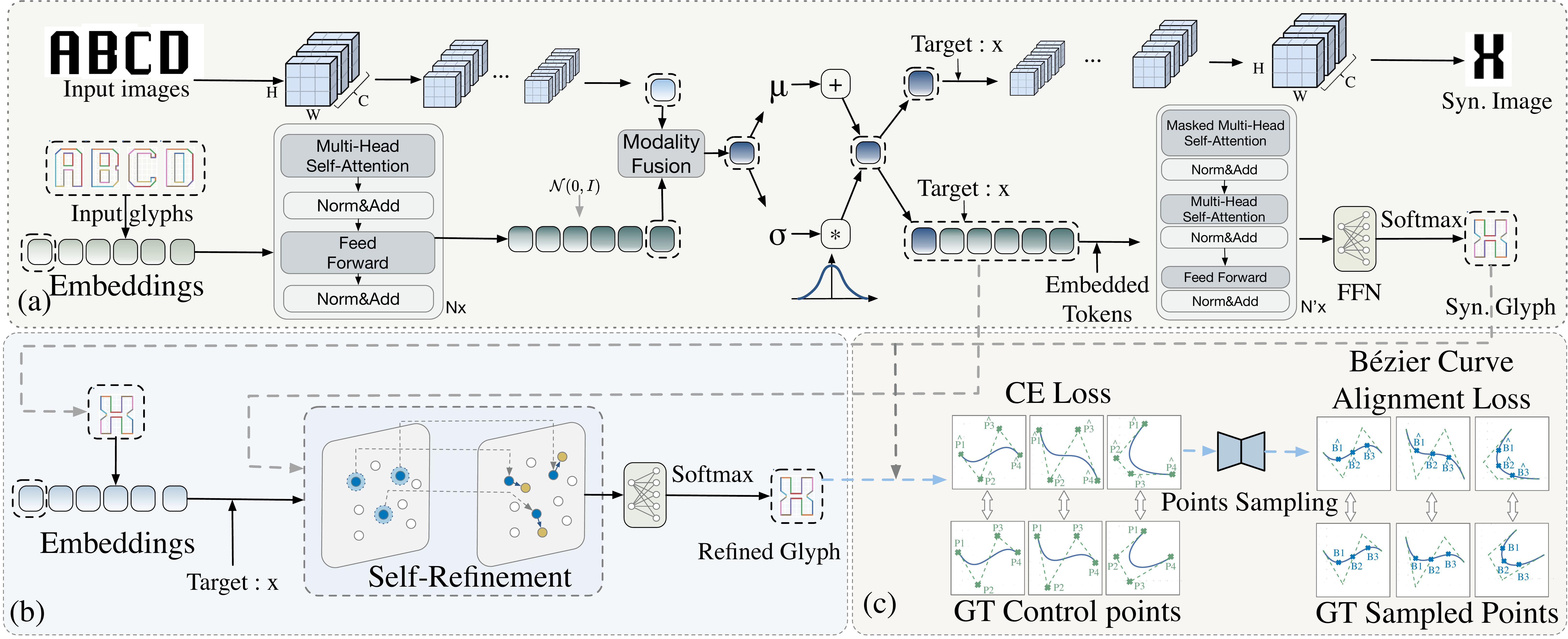}
\end{center}
\vspace{-0.3cm}
   \caption{The pipeline of our DeepVecFont-v2. The inputs are reference glyphs in both raster images and vector outlines. (a) A dual-branch architecture based on Transformers and CNNs aims to synthesize the target vector glyph. (b) The self-refinement module is designed to remove artifacts in the initially synthesized vector glyphs. (c) In addition to control points, auxiliary points are sampled to align the synthesized glyph with the corresponding target via the Bézier curve alignment loss.
   }

\label{architecture}
\end{figure*}
    
\begin{figure}[t!]
    \begin{center}
    \includegraphics[width=\columnwidth]{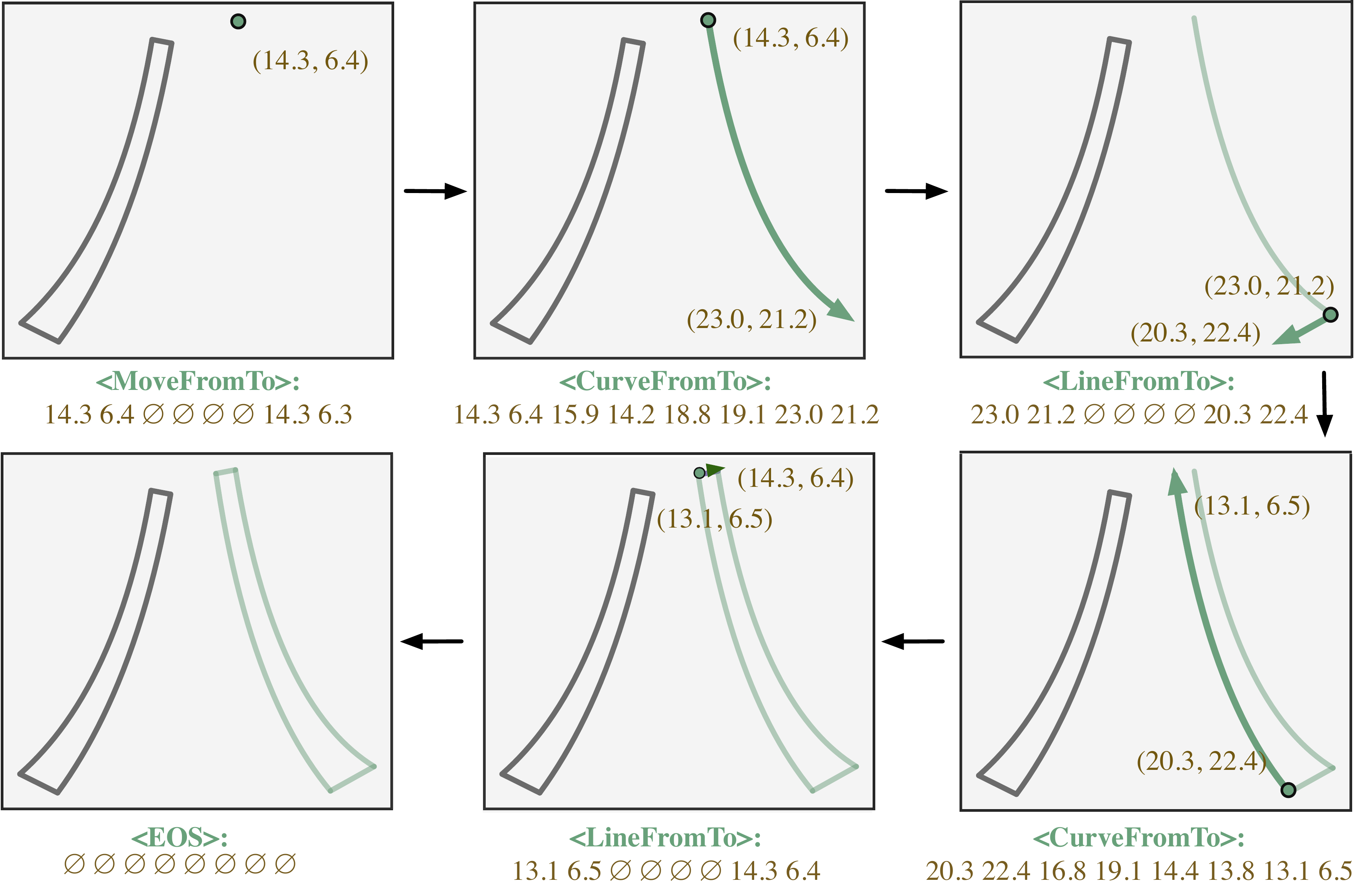}
    \end{center}
       \caption{An illustration of our data structure used to describe vector glyphs. The command type and coordinates are shown below each canvas. The origin of the axes is at the top left of each canvas and ``$\emptyset$" denotes the unused argument.}
    
    \label{data_struc_ilus}
    \end{figure}

\section{Method}
In this section, we present the details of our proposed DeepVecFont-v2. Specifically, we first introduce the data structure and the learned SVG embeddings of vector glyphs. Then, we describe the overview of our network architecture and the implementation details of each module.

\subsection{Data Structure and SVG Embeddings}
A vector font is a set of vector glyphs $\{G_1,...,G_{Nchar}\}$, where $N_{char}$ denotes the number of character categories. As shown in Fig.~\ref{data_struc_ilus}, a vector glyph $G_i$ can be viewed as a sequence of drawing commands, denoted as $G_i =[C_{i,1},..., C_{i,N_c}]$, where $N_c$ is the total length of commands. The drawing command $C_{i,j}$ is a tuple $(z_{i,j},p_{i,j})$, where $z_{i,j}$ and $p_{i,j}$ denote the command type and command coordinates, respectively.
The drawing sequence starts from the top-left command.
We consider 4 command types, i.e., $z_i \in \{MoveFromTo, LineFromTo, CurveFromTo, EOS\}$. $MoveFromTo$ means moving the drawing position to a new location, which is used for starting a new path. $LineFromTo$ and $CurveFromTo$ mean drawing a straight line and a three-order Bézier curve, respectively. $EOS$ means ending the drawing sequence.\par
Typically, 
$p_{i,j}$ is made up of $N_p$ pairs of coordinates: $p_{i,j}=[(x_{i,j}^1,y_{i,j}^1),...,(x_{i,j}^{N_p},y_{i,j}^{N_p})]$, known as the control points, where $N_p$ is determined by the order of Bézier curves. In the typical SVG representation, the starting point is omitted so that the number of curve order is equal to $N_p$, namely, $N_p$ = 3 for the three-order Bézier curve and $N_p$ = 1 for the line (equal to the one-order Bézier curve). \par
\textbf{Relaxation Representation} Different from SVG-VAE and DeepVecFont, we assign each drawing command with individual starting and ending points.
Therefore, the number of coordinate pairs $N_p$ in $CurveFromTo$ is set to 4.
Specifically, $p_{i, j}^1$ and $p_{i, j}^4$ are the starting and ending points, respectively; $p_{i, j}^2$ and $p_{i, j}^3$ are two intermediate control points. 
We pad the length of other commands to that of $CurveFromTo$: for $z_{i,j} \in \{MoveFromTo, LineFromTo\}$, only $(x_{i,j}^1,y_{i,j}^1) $ and $(x_{i,j}^4,y_{i,j}^4)$ (starting and ending points) are used; for $z_{i,j} = EOS$, no argument is used. Afterwards, we render those vector glyphs to obtain the rasterized glyph images.

\begin{figure}[t!]
    \begin{center}
    \includegraphics[width=\columnwidth]{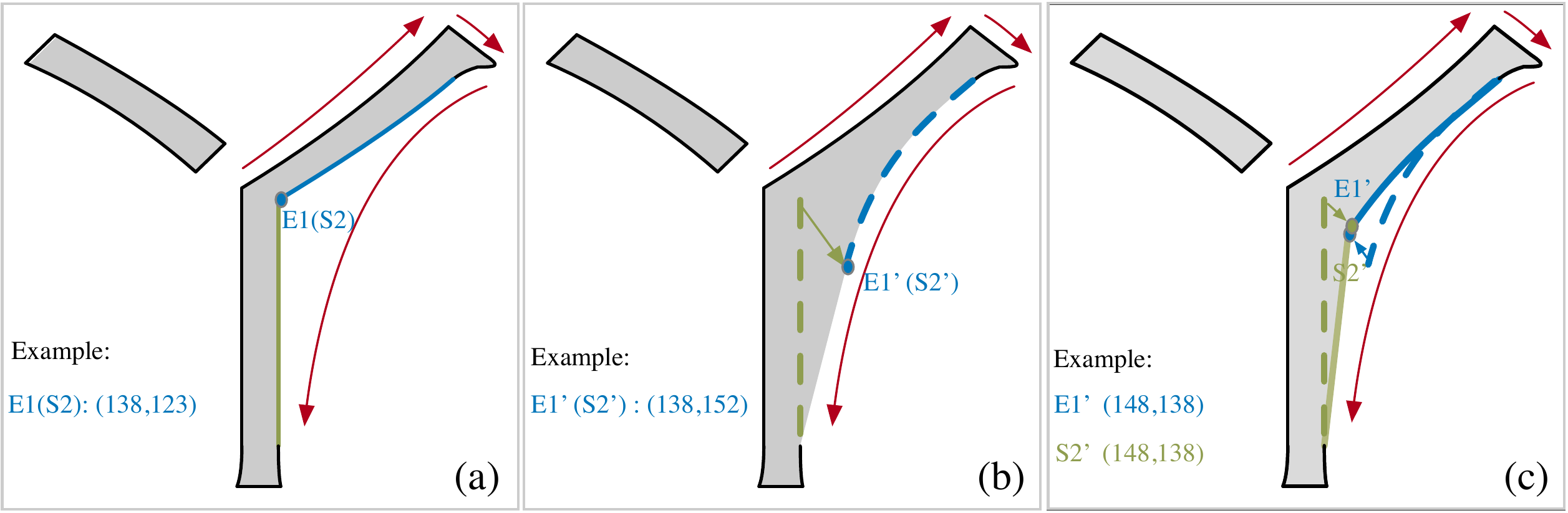}
    \end{center}
        \caption{A demonstration of different SVG representations. (a) The ground truth. If the previous blue segment is wrongly predicted, (b) the commonly used representation in SVG-VAE \cite{svgvae} shares the connected points (E1 and S2), resulting in location shift. (c) The proposed relaxation representation models these two points separately, which is more robust against outliers.}

    
    \label{svg_rep_comp}
    \end{figure}

 An illustration of the difference between an existing SVG representation and ours can be found in Fig.~\ref{svg_rep_comp}. Our proposed representation describes each drawing command separately, which is well-suited for the Transformer's attention mechanism to exploit the long-range dependencies of drawing commands. In the training stage, we force the ending point of $C_i$ to be consistent with the starting point of $C_{i+1}$ using an extra constraint calculated by Eq.~\ref{smoothconstraint}. In the inference stage, we merge these two points by averaging their positions.

    \textbf{Embedding}
We first project the drawing command into a common continuous $d_E$-dimensional embedding space. 
Specifically, each $C_{i,j}$ is projected into a vector $e_{i,j} \in \mathbb{R}^{d_{E}}$ via the sum of four embeddings: 
\begin{equation}
e_{i,j}=e_{i,j}^{\mathrm{cmd}}+e_{i,j}^{\mathrm{args}}+ e_{i,j}^{\mathrm{w,h}} + e_{i,j}^{\mathrm{pos}}. 
\end{equation}

For the command type embedding $e_{i,j}^{\mathrm{cmd}}$, we use a learnable matrix $W_{\mathrm{cmd}} \in \mathbb{R}^{d_{E} \times 4}$ to convert the command type into a $d_E$-dimensional vector, which is formulated as $e_{i,j}^{\mathrm{cmd}}=W_{\mathrm{cmd}} \delta_{i,j}^c \in \mathbb{R}^{d_{E}}$, where $\delta_{i,j}^c \in \mathbb{R}^{4}$ is a one-hot vector for the four command types.

For the argument (coordinate) embedding $e_{i,j}^{\mathrm{args}}$, we first quantize the continuous coordinates into discrete integers and convert the integer into a one-hot vector with 256 dimensions. Then, we stack all the 8 coordinate parameters into a matrix ${\delta}_{i,j}^{\mathrm{p}} \in \mathbb{R}^{256 \times 8}$, and embed each parameter using a learnable matrix ${W}_{{args}}^{b} \in \mathbb{R}^{d_{\mathrm{E}} \times 256}$. After that, we aggregate all the parameter embeddings through a linear projection layer ${W}_{{args}}^{a} \in \mathbb{R}^{d_{\mathrm{E}} \times 8 d_{\mathrm{E}}}$, which is formulated as:

\begin{equation}
    e_{i,j}^{{args}}=W_{ {args}}^{a} { flatten }\left(W_{ {args}}^{b} {\delta}_{i,j}^{\mathrm{p}}\right),
    \end{equation}
where flatten(·) means flattening the input into a vector.

The third term $e_{i,j}^{\mathrm{w,h}}$ encodes the height and width of the glyph area to capture global styles. We discretize the height and width into discrete integers and project them into the continuous space to obtain $e_{i,j}^{\mathrm{w}}$ and $e_{i,j}^{\mathrm{h}}$. Then we concatenate them by element-wise addition to get $e_{i,j}^{\mathrm{w,h}}$.

The fourth term (positional embedding) $e_{i,j}^{\mathrm{pos}}$ encodes the position and order information of all commands in a drawing sequence. Similar to \cite{2017Attentionisallyouneed}, we use the absolute positional encoding to compute the $e_{i,j}^{\mathrm{pos}} \in \mathbb{R}^{d_{E}} $ for each command.

\subsection{Dual-branch Pipeline}
Fig.~\ref{architecture} shows the pipeline of our DeepVecFont-v2. Given randomly sampled $N_r$ (typically set to 4) reference glyphs as input, the model generates a target glyph with the same font style as input samples, which is further refined by a self-refinement module. Similar to DeepVecFont~\cite{wang2021deepvecfont}, the proposed model also adopts a dual-branch architecture which, however, employs several new techniques and modules. More details are presented as follows:


\textbf{Encoder}
The encoder of our DeepVecFont-v2 is made up with an image encoder and a sequence encoder. The image encoder is a CNN, outputting the image feature $f_{img} \in \mathbb{R}^{d_{E}}$. Different from DeepVecFont, the sequence encoder is a Transformer encoder composed of six layers of Transformer encoder blocks. 
For each vector glyph $G_i$, it receives as input the sequence embedding $[e_{i,0},e_{i,1},...,e_{i,N_c}]$, where $[e_{i,1},...,e_{i,N_c}]$ is the embedding of $N_c$ drawing commands and $e_{i,0}$ denotes an auxiliary learnable ``token" for performing dual-modality fusion.
The output of our Transformer encoder is denoted as $e_i^{\prime} = [e_{i,0}^{\prime},e_{i,1}^{\prime},...,e_{i,N_c}^{\prime}] \in \mathbb{R}^{d_{E} \times (N_{c+1})}$. Next, we calculate the holistic sequence-aspect style feature $f_{seq} = [f_0^{seq},f_1^{seq},...,f_{N_c}^{seq}]$, where $f_j^{seq}$ is aggregated from all the $j$-th tokens in $e^{\prime}$ of $N_{r}$ reference glyphs via linear projection.



\textbf{Modality fusion}
We use the modality token ${f_0^{seq}}$ to represent the style feature of vector glyphs, and combine it with the image feature $f_{img}$ by a linear projection layer to get the fused feature $f$:

\begin{equation}
    f=Linear\left(\left[f^{i m g} ; f_0^{seq}\right]\right),
    \end{equation}
then $f$ is normalized by using the reparametrization trick introduced in VAE~\cite{2014Autoencoder}.

\textbf{Decoder}
The image decoder is a Deconvolutional Neural Network. We send $f$ into the image decoder to generate the target glyph image $I_t$. Then we employ the L1 loss and perceptual loss to compute the image reconstruction loss:

\begin{equation}
    L_{img}=\left\|\hat{I}_{t}-I_{t}\right\|_{1}+L_{\text {percep }}\left(\hat{I}_{t}, I_{t}\right).
\end{equation}

We feed the sequence decoder with the input of $[f, f_1^{seq},..., f_{N_c}^{seq}]$, where the original modality token $f_0^{seq}$ is replaced with the fused feature $f$.
An MLP layer is appended on top of the Transformer decoder, predicting the command types and coordinates of the target glyph as $[\hat{z}_{t,1},...\hat{z}_{t,N_c}]$ and $[\hat{p}_{t,1},...\hat{p}_{t,N_c}]$, respectively. Here we define the loss between the initially generated glyph and its corresponding ground truth as:

    \begin{equation}
        L_{CE}^{init}=\sum_{j=1}^{N_{C}}w_{\mathrm{cmd}} \ell\left(z_{t,j}, \hat{z}_{t,j}\right)+ 
        \ell \left(p_{t,j},\hat{p}_{t,j}\right),
        \end{equation}
    where $\ell$ denotes the Cross-Entropy loss, all the coordinates ($p$) are quantized to be optimized by $\ell$, $w_{\mathrm{cmd}}$ means the loss weight for command type prediction, and all the unused arguments are masked out in the second item.

  \subsection{Context-based Self-refinement}
    In the inference stage, Transformers follow the sequential generation process adopted in RNNs: taking the previous predicted results as the condition to predict the following steps. 
    However, a prediction error generated in a certain step of the sequential process might finally lead to the accumulation of large errors and bring significant distortions in the synthesized glyphs. We notice that there are strong priors and correlations in the geometries of glyphs, such as symmetry and smoothly varied stroke widths, 
    and these context information can be used as strong clues to enhance the glyph synthesizing performance.
    \par
    Therefore, we propose a self-refinement module to refine the initial predictions by analyzing their context information, i.e., $((\bar{z}_{t,1},\bar{p}_{t,1}), \ldots, (\bar{z}_{t,N_c},\bar{p}_{t,N_c})) = F_{r}((\hat{z}_{t,1},\hat{p}_{t,1}), \ldots, (\hat{z}_{t,N_c},\hat{p}_{t,N_c}))$, where $F_{r}$ denotes the function of our refinement module which is actually a 2-layer Transformer decoder. The multi-head attention operation ($\mathbf{F}_{mha}$) used in our decoder layers is formulated as:
    
    \begin{equation}
        \mathbf{F}_{m h a}=\operatorname{softmax}\left(\frac{\mathbf{Q K}^{\top}}{\mathcal{C}(\mathbf{Q})}+\mathbf{M}\right) \mathbf{V},
    \end{equation}
    \begin{equation}
        \mathbf{M}_{i j}= \begin{cases}0, & i <=\hat{N}_c \\ -\infty, & i>\hat{N}_c\end{cases},
    \end{equation}
    where $\hat{N}_c$ is the length of the predicted sequence; 
    $\mathcal{C}$ is the feature dimension of queries for normalization; $\mathbf{Q} \in \mathbb{R}^{d_E \times N_c}$ consists of the embeddings of the initially predicted sequence $(\hat{z}_{t,1},\hat{p}_{t,1}), \ldots, (\hat{z}_{t,N_c},\hat{p}_{t,N_c})$; $\mathbf{K,V} \in \mathbb{R}^{d_{E} \times\left(N_{c}+1\right)}$ denote the linear projections of memories used in the previous sequence decoder; $\mathbf{M}$ is used to perform self-attention only on the valid positions of the initially predicted sequence. Finally, we refine the command types and coordinates $(\bar{z}_{t,j},\bar{p}_{t,j})$ again via the supervision of the corresponding ground truth by minimizing:
    
    \begin{equation}
        L_{CE}^{refine}=\sum_{j=1}^{N_{C}}\left(w_{\mathrm{cmd}} \ell\left(z_{t,j}, \bar{z}_{t,j}\right)+ \ell \left(p_{t,j},\bar{p}_{t,j}\right)\right). 
        \end{equation}

\subsection{Bézier Curve Alignment Loss}
    Through our experiments, we found that it is still insufficient by only employing the control points ($p$) to supervise the alignment of Bézier curves/lines.
    Therefore, we propose to sample more points from each drawing command $C_{t,j}$. 
    As we know, a Bézier curve built upon control points $p$ is defined as:
    $B=\sum_{k=0}^{n} 
        \left(\begin{array}{c}
        n \\
        k
        \end{array}\right)  r^{k}(1-r)^{n-k}p,$
where $n$ denotes the number of curve order,
        $\left(\begin{array}{c}
        n \\
        k
        \end{array}\right)$ denotes the binomial coefficients,
 and $r$ is the position parameter.
    Thus, we can calculate the Bézier curve alignment loss by:
    \begin{equation}
         L_{b\Acute{e}zier} = \sum_{r\in r_p} (\hat{B}_{t,j}(r)- B_{t,j}(r))^2 + (\bar{B}_{t,j}(r)- B_{t,j}(r))^2,
    \end{equation}
    where $r_{p} = \{0.25, 0.5, 0.75\}$ represents the parameters of the auxiliary points we sampled.
Note that ``LineFromTo" can be formulated as a special case of ``CurveFromTo", where all the control points are uniformly distributed on a line segment.

\subsection{Merging Relaxed Drawing Commands}
Ideally, the starting point of the current drawing command ($LineFromTo$ or $CurveFromTo$, $l$ or $c$ for short) should coincide with the previous one's ending point for both initially synthesized and refined sequences. Thereby, we try to minimize the L2 distance between the predicted ending point of command $C_{i-1}$ and the starting point of $C_{i}$:
\begin{equation}
\begin{aligned}
    L_{cons} &=\mathbf{1}_{z_{t,j}\in\{l,c\}}\sum_{j=2}^{N_c}\left(\hat{x}_{t,j}^{1}-\hat{x}_{t,j-1}^{4}\right)^2 + \left(\hat{y}_{t,j}^{1}-\hat{y}_{t,j-1}^{4}\right)^2 \\
&+ \left(\bar{x}_{t,j}^{1}-\bar{x}_{t,j-1}^{4}\right)^2 + \left(\bar{y}_{t,j}^{1}-\bar{y}_{t,j-1}^{4}\right)^2.
    \label{smoothconstraint}
\end{aligned}
\end{equation}

\subsection{Overall Objective Loss}
Combining all losses mentioned above, we train the whole model by minimizing the following objective:
\begin{equation}
  L_{img} + L_{CE}^{init} + L_{CE}^{refine} + L_{cons} + L_{b\Acute{e}zier} +L_{kl},
  \label{TotalLoss}
\end{equation}
where the last term represents the KL loss normalizing the latent code $f$. For brevity, the weight of each term is omitted from the equation.

        
        \begin{figure}[t!]
    \begin{center}

    \includegraphics[width=\columnwidth]{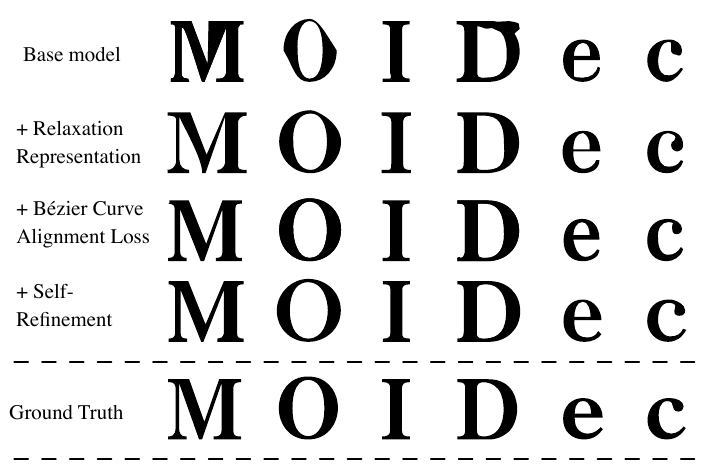}
    \end{center}
      \vspace{-0.5cm}
        \caption{The ablation study of our method. Blue circles indicate the shortcomings of each incomplete configuration.}
    \label{ablations-en}
    \end{figure}

        \begin{figure*}[t!]
        \begin{center}
        \includegraphics[width=\textwidth]{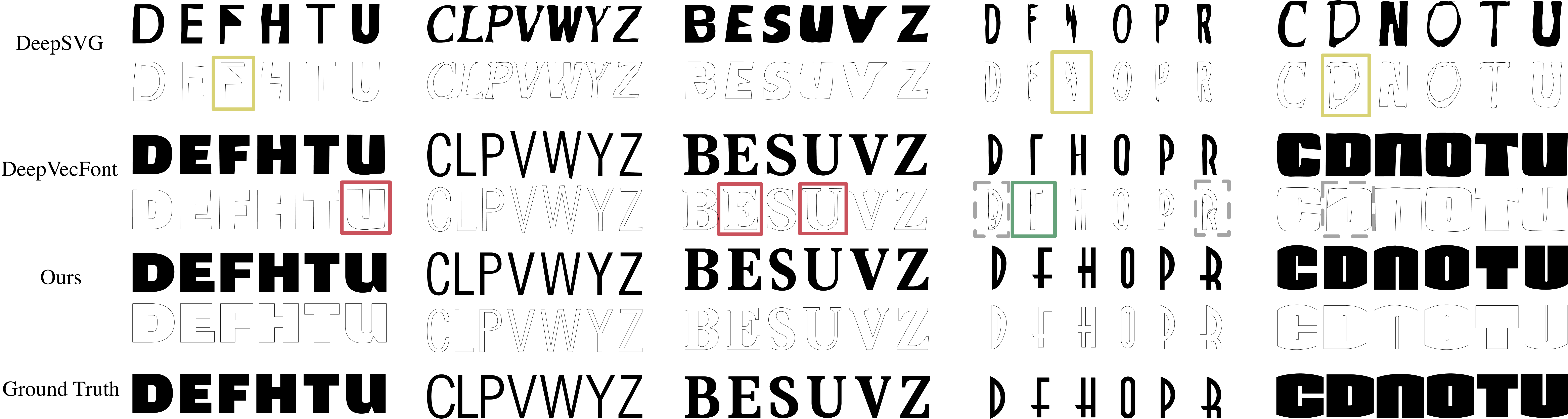}
        \end{center}
          \vspace{-0.4cm}
           \caption{Comparison of few-shot font generation results for DeepSVG, DeepVecFont, and our method. Our method shows superiority in generating high-quality English vector fonts with various styles. Please zoom in for better inspection.} 
        
        \label{SOTA-en}
        \end{figure*}     

        \section{Experiments}
        \subsection{Dataset and Implementation Details}
        
        The dataset used in our experiment for generating English fonts is the same as~\cite{wang2021deepvecfont}, which includes 8035 fonts for training and 1425 fonts for testing. For Chinese font synthesis, we built a dataset consisting of 212 and 34 fonts for training and testing. To directly apply the same model in two datasets, here we only randomly selected 52 Chinese characters with relatively simple structures. Since the scale of our dataset for Chinese fonts is relatively small, we augmented the training set by applying the affine transformation, enlarging it by ten times. 
        
        We conduct the experiments by following the experimental settings of DeepVecFont~\cite{wang2021deepvecfont} for fair comparison. Specifically, the numbers of reference glyphs are chosen as 4 and 8 for English and Chinese font generation, respectively. We employ the Adam optimizer with an initial learning rate of 0.0002. The image resolution is chosen as $64\times64$ in both the training and testing stages. When inferencing, we first add a noise vector distributed by $\mathcal{N}(0, I)$ to the sequence feature, to simulate the feature distortion caused by the human-designing uncertainty (as observed in DeepVecFont). Then, we sample $N_s$ (10 for English and 50 for Chinese) synthesized vector glyphs as candidates and select the one as the final output that has the highest IOU value with the synthesized image. The reconstruction errors (denoted as ``Error") in our quantitative results are obtained by computing the average L1 distance between the rasterized image of each synthesized vector glyph and its corresponding ground-truth glyph image (at the resolution of $64\times64$) in the testing set. In our experiments, the weight values of different losses in Eq.\ref{TotalLoss} are set to 1.0, 1.0, 1.0, 10, 1.0, and 0.01, respectively from left to right.

        \subsection{Ablation Study}
        
        \begin{figure}[t!]
        \vspace{-0.5cm}
            \begin{center}
                
                \includegraphics[width=\columnwidth]{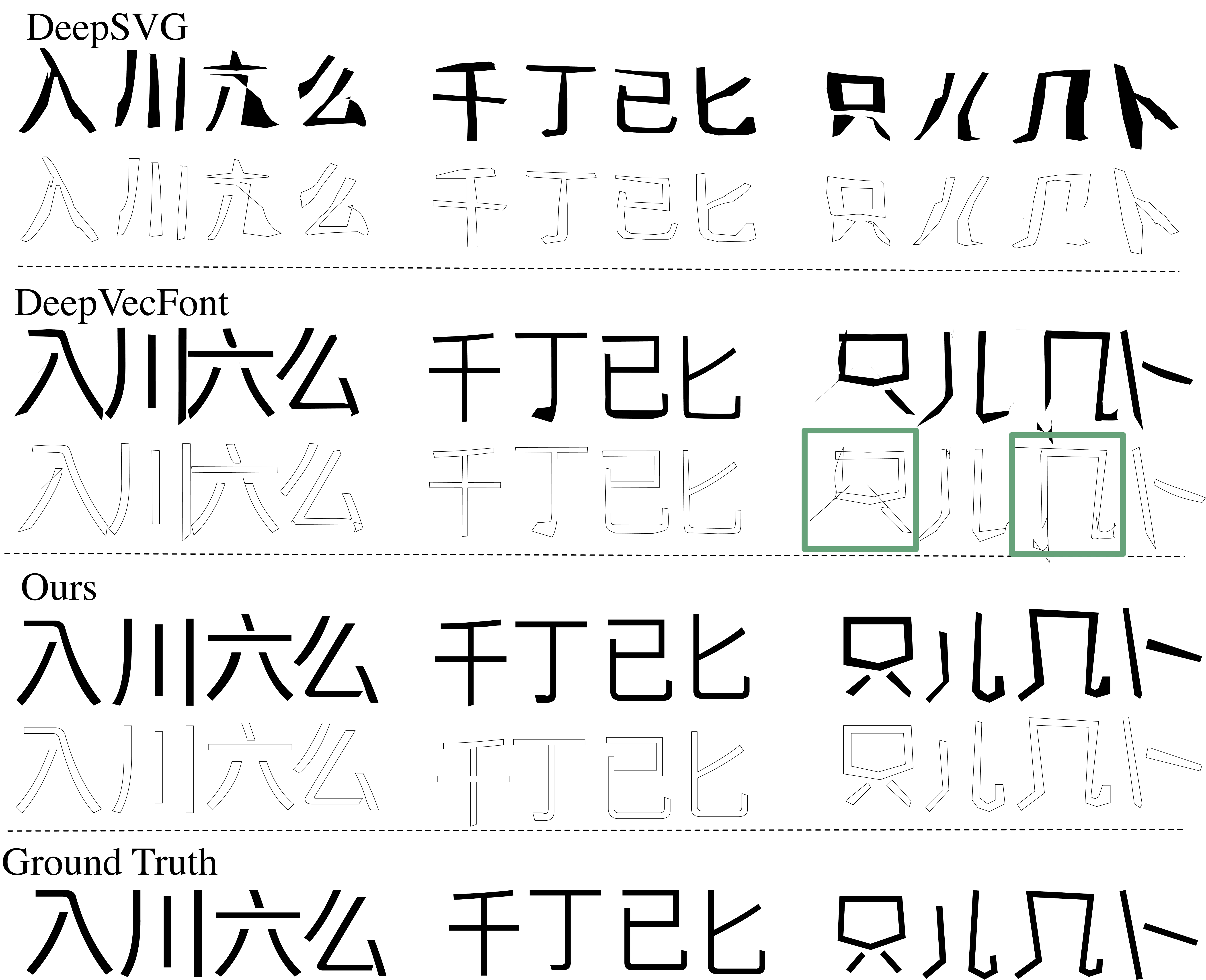}
                \end{center}
                  \vspace{-0.5cm}
                    \caption{Comparison of few-shot Chinese vector font generation results obtained by different methods.
                    }
                
                \label{SOTA-CN}
                \end{figure}
        We conduct qualitative and quantitative experiments to examine the impact of each module in our proposed model. In Tab.~\ref{ablations-en}, the base model is modified from DeepVecFont by simply replacing the LSTM networks with Transformers, then we evaluate each module by adding them successively to the base model. As shown in Fig.~\ref{ablations-en}, the base model using the SVG representation in SVG-VAE~\cite{svgvae} tends to generate incomplete glyphs with severe distortions. By employing the relaxation representation, semantic information can be preserved and the smooth connection between adjacent drawing commands is guaranteed. However, there still exist suboptimal curves with slight distortions. After adding the Bézier curve alignment loss to the current model, most distortions on the synthesized curves can be eliminated. Finally, we perform self-refinement to remove artifacts in the initially synthesized glyphs, resulting in high-quality glyphs with compact and coordinated outlines. Tab.~\ref{table-ablation} shows some quantitative results of our ablation study, further demonstrating the superiority of our proposed modules.

        \subsection{Parameter Study}
        We also conduct parameter studies to find the best choice of the number of sampling points distributed along the Bézier curves, and the results are shown in Tab.\ref{NumberOfSampledPoints}. we can see that the model's performance will be markedly improved when the number of sampling points changes from 0 to 3, while the performance is just slightly enhanced when we increase the point number from 3 to 9. This is mainly due to the fact that 3 sampling points are sufficient enough to precisely align two Bézier curves/lines. Therefore, we choose to sample 3 points for each Bézier curve/line to achieve a balance between the effectiveness and efficiency of our method. Furthermore, we also conduct parameter studies to examine the performance of our method under different numbers of input reference glyphs. Please refer to the supplemental materials for more details. 

         \begin{table}[t!]
        \centering
        \begin{tabular}{lc}
        \hline
        Model &  Error-EN$\downarrow$ \\
        \hline
        Base model &   0.0588  \\
        + Relaxation Representation &  0.0557 \\
        + Bézier Curve Alignment Loss &  0.0529 \\
        + Self-Refinement &   0.0519 \\
        \hline
        \end{tabular}
        \caption{Comparison of reconstruction errors for our methods under different configurations. }
        \label{table-ablation}
        \end{table}   

        \begin{figure}[t!]
            \begin{center}
            
              
                \includegraphics[width=\columnwidth]{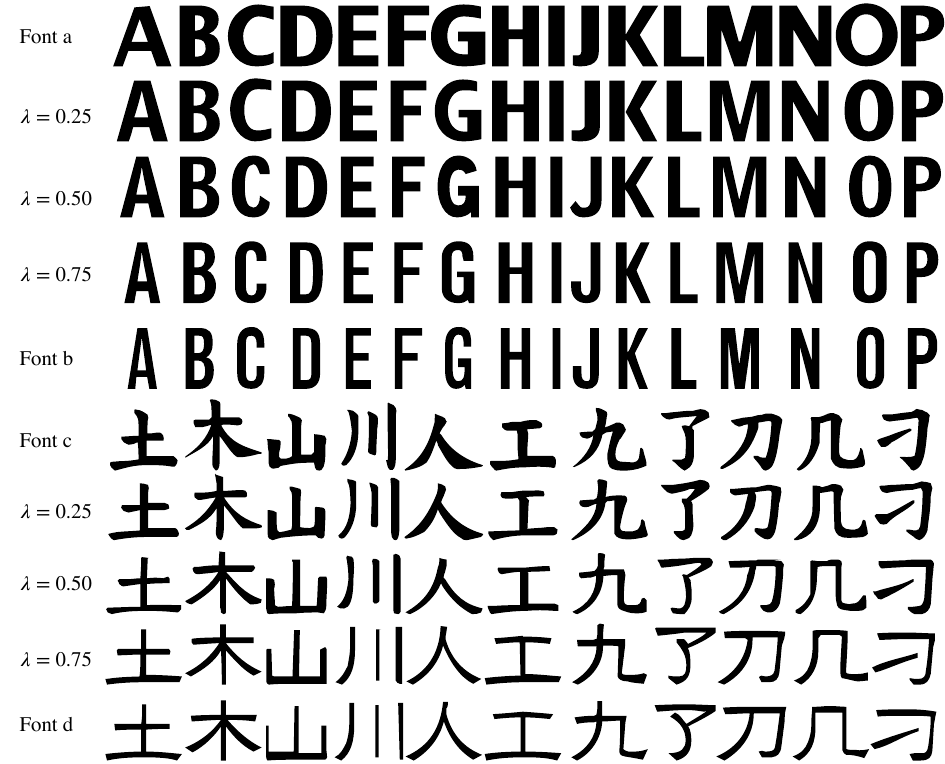}
                \end{center}
                  \vspace{-0.3cm}
                    \caption{English and Chinese vector font interpolation results, where the weight, width and styles change smoothly.}
                
                \label{interpolation}
                \end{figure}

  \begin{figure*}[t!]
    \begin{center}
    \hspace{-0.5cm}
    \includegraphics[width=\textwidth]{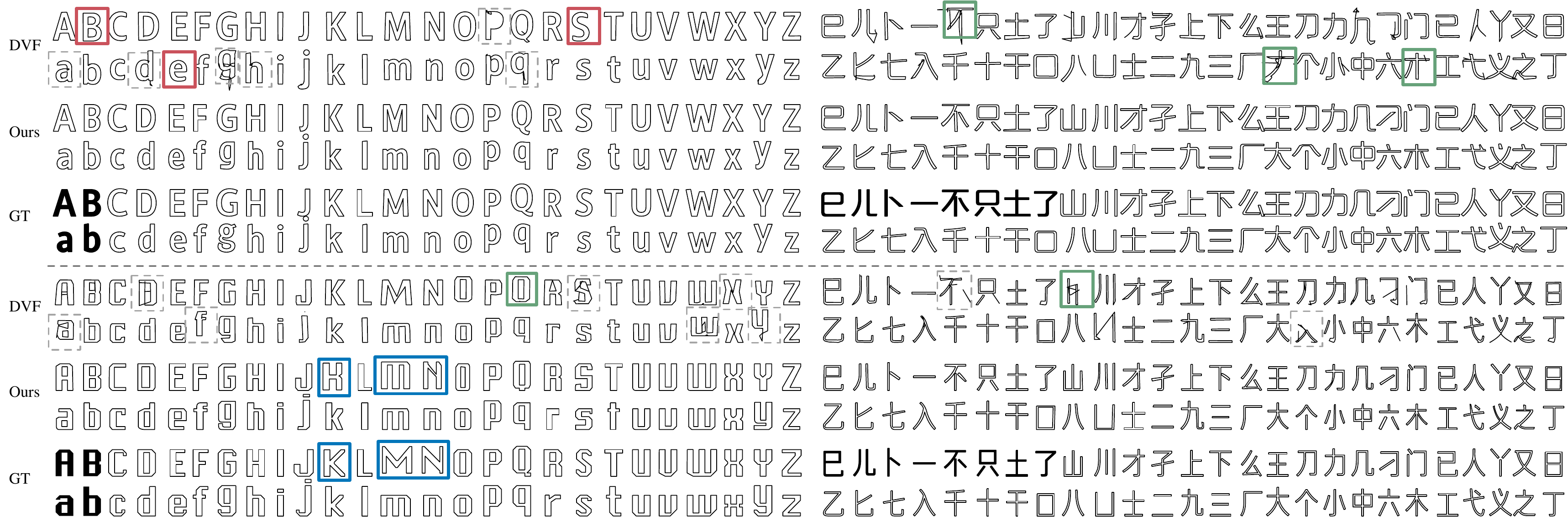}
    \end{center}
    \vspace{-0.5cm}
       \caption{Comparison of more vector fonts synthesized by DeepVecFont (DVF) and our DeepVecFont-v2 in the task of few-shot font generation. The input reference glyphs are filled in black. Please zoom in for better inspection.} 
    \label{all-en-cn}

    \end{figure*}    

        \subsection{Font Interpolation}
        Our method can be used to achieve smooth interpolation between two vector fonts. Specifically, given two vector fonts $a$ and $b$, an interpolated style feature between them can be calculated by:
        \begin{equation}
            f_{inter}=(1-\lambda) \cdot f(a)+\lambda \cdot f(b).
            \end{equation}
        Then, we feed the interpolated feature into the sequence and image decoders to generate the interpolated vector glyphs and the corresponding glyph images. Fig.~\ref{interpolation} shows two font interpolation results for English and Chinese vector fonts, respectively. We can see that glyphs in the source styles can smoothly morph to the target ones, demonstrating the effectiveness of our method in vector font interpolation.

        \begin{table}[t!]
        \centering
        \begin{tabular}{lcc}
        \hline
        Model &  Error-EN$\downarrow$ & Error-CN$\downarrow$ \\
        \hline
        DeepSVG & 0.125  &    0.167  \\
        DeepVecFont & 0.056 &  0.086  \\
        Ours &  0.052   &   0.080  \\
        \hline
        \end{tabular}
          \vspace{-0.1cm}
        \caption{Comparison of reconstruction errors for different methods. ``EN" and ``CN" denote the English and Chinese testing datasets, respectively.}
        \label{table-dvf}
        \end{table}

                 \begin{table}[t!]
        \centering
        \begin{tabular}{ccccc}
        \hline
        Point Num&  Error-EN$\downarrow$&Point Num &  Error-EN$\downarrow$ \\
        \hline
        0 &  0.0557  & 6 &   0.0526  \\ 
        1 &    0.0543  &9 &     0.0524   \\ 
        3 &   0.0529 & 12 &  0.0520\\  
        
        \hline
        \end{tabular}
          \vspace{-0.1cm}
        \caption{Comparison of reconstruction errors under different numbers of sampling points. ``EN" denotes the English testing dataset.}
        \label{NumberOfSampledPoints}
        \end{table}
        

   \subsection{Comparison with the State of the Arts.}
   \vspace{-0.2cm}
We compare the performance of our DeepVecFont-v2 with other existing methods (i.e., DeepSVG~\cite{carlier2020deepsvg} and DeepVecFont\cite{wang2021deepvecfont}) on English and Chinese vector font datasets. Some qualitative results are shown in Fig.~\ref{SOTA-en}, Fig.~\ref{SOTA-CN}, and Fig.~\ref{all-en-cn}. The quantitative results are shown in Tab. \ref{table-dvf}.

\textbf{English vector font generation.}
From Fig. \ref{SOTA-en}, we can see that the vector glyphs synthesized by DeepSVG often contain severe distortions (marked in yellow). This is mainly because the hierarchical architecture of DeepSVG was originally designed for synthesizing vector icons with multiple separated short-length paths, while a vector glyph typically consists of several long-range outlines. The synthesized glyphs of DeepVecFont filled with black pixels generally look visually pleasing. However, there still exist some problems in the synthesis results regarding the details of outlines: 1) the refinement post-processing tends to over-fit the initially synthesized vector glyph with the corresponding raster image, resulting in suboptimal outlines with over-smoothed corners (marked in red); 2) when a sequence with redundant drawing commands is predicted, the outline tends to be staked together with self-interactions (marked in gray); 3) structurally-incorrect glyphs might be synthesized for some special styles (e.g., ``F" in the green box). On the contrary, our method can synthesize high-quality vector glyphs for almost all kinds of font styles. Fig. \ref{all-en-cn} shows more results obtained by our method and DeepVecFont. From Fig. \ref{all-en-cn}, we can also observe that our DeepVecFont-v2 sometimes can even generate vector glyphs whose font styles are more consistent with the input samples compared to the ground truth (marked in blue). 

        
\textbf{Chinese vector font generation.}
Chinese glyphs typically contain complex shapes and structures, making the task of Chinese vector font synthesis much more challenging. As shown in Fig. \ref{SOTA-CN}, DeepSVG tends to generate glyphs with inconsistent styles and severe artifacts, and DeepVecFont may obtain fragmented results when synthesizing glyphs with multiple closed paths (marked in green). In contrast, our method synthesizes visually pleasing vector fonts, mainly due to the utilization of Transformers and our specifically-designed modules to guarantee the smoothness of outlines. Fig. \ref{all-en-cn} shows more results obtained by our method and DeepVecFont. The quantitative results shown in Tab. \ref{table-dvf} further demonstrate the superiority of our method when handling fonts consisting of glyphs with more complex topologies and longer drawing-command sequences. More synthesized results can be found in our supplemental materials.

\vspace{-0.2cm}
\subsection{Limitations}
   \vspace{-0.2cm}
Our model fails to synthesize correct results when handling complex glyphs that contain large amounts of long drawing paths (e.g., ``\begin{CJK*}{UTF8}{gbsn}霸\end{CJK*}" shown in our supplemental materials). One possible reason is that longer sequences inherently exist more uncertainties generated during the human font-designing process~\cite{wang2021deepvecfont}, bringing more challenges to train the model. 
Another reason is the possible existence of complicated topological changes for a Chinese character in different styles, making it hard to learn stable SVG embeddings. We leave these issues as our future work.

\vspace{-0.2cm}
\section{Conclusion}
   \vspace{-0.2cm}
This paper proposed a novel method, DeepVecFont-v2, to effectively handle the challenging task of vector font synthesis. Specifically, we replaced RNNs adopted in the original DeepVecFont by Transformers and adopted several specifically-design modules, including relaxation representation, the Bézier curve alignment loss, and context-based self-refinement. Thus, vector fonts with higher quality can be directly synthesized by our method in an end-to-end manner. Experimental results demonstrated the superiority of our DeepVecFont-v2 compared to the state of the art in the applications of English and Chinese vector font synthesis.

\section*{Acknowledgements}
This work was supported by National Language Committee of China (Grant No.: ZDI135-130), Project 2020BD020 supported by PKU-Baidu Fund, Center For Chinese Font Design and Research, and Key Laboratory of Science, Technology and Standard in Press Industry (Key Laboratory of Intelligent Press Media Technology).

\clearpage
{\small
\bibliographystyle{ieee_fullname}
\bibliography{egbib}

\begin{thebibliography}{10}\itemsep=-1pt

\bibitem{svgforcn}
Haruka Aoki and Kiyoharu Aizawa.
\newblock Svg vector font generation for chinese characters with transformer.
\newblock {\em arXiv preprint arXiv:2206.10329}, 2022.

\bibitem{azadi2018multi}
Samaneh Azadi, Matthew Fisher, Vladimir~G Kim, Zhaowen Wang, Eli Shechtman, and
  Trevor Darrell.
\newblock Multi-content gan for few-shot font style transfer.
\newblock In {\em Proceedings of the IEEE conference on computer vision and
  pattern recognition}, pages 7564--7573, 2018.

\bibitem{1994Mixture}
C.~M. Bishop.
\newblock Mixture density networks.
\newblock {\em IEEE Computer Society}, 1994.

\bibitem{fontmf}
Neill~DF Campbell and Jan Kautz.
\newblock Learning a manifold of fonts.
\newblock {\em ACM Transactions on Graphics (ToG)}, 33(4):1--11, 2014.

\bibitem{carlier2020deepsvg}
Alexandre Carlier, Martin Danelljan, Alexandre Alahi, and Radu Timofte.
\newblock Deepsvg: A hierarchical generative network for vector graphics
  animation.
\newblock {\em Advances in Neural Information Processing Systems},
  33:16351--16361, 2020.

\bibitem{gao2019agisnet}
Yue Gao, Yuan Guo, Zhouhui Lian, Yingmin Tang, and Jianguo Xiao.
\newblock Artistic glyph image synthesis via one-stage few-shot learning.
\newblock {\em ACM Transactions on Graphics (TOG)}, 38(6):1--12, 2019.

\bibitem{2014Generative}
Ian Goodfellow, Jean Pouget-Abadie, Mehdi Mirza, Bing Xu, David Warde-Farley,
  Sherjil Ozair, Aaron Courville, and Y. Bengio.
\newblock Generative adversarial nets.
\newblock In {\em Neural Information Processing Systems}, 2014.

\bibitem{ha2017sketchrnn}
David Ha and Douglas Eck.
\newblock A neural representation of sketch drawings.
\newblock {\em arXiv preprint arXiv:1704.03477}, 2017.

\bibitem{1997Lstm}
S. Hochreiter and J. Schmidhuber.
\newblock Long short-term memory.
\newblock {\em Neural Computation}, 9(8):1735--1780, 1997.

\bibitem{adain}
Xun Huang and Serge Belongie.
\newblock Arbitrary style transfer in real-time with adaptive instance
  normalization.
\newblock In {\em Proceedings of the IEEE international conference on computer
  vision}, pages 1501--1510, 2017.

\bibitem{pix2pix}
Phillip Isola, Jun-Yan Zhu, Tinghui Zhou, and Alexei~A Efros.
\newblock Image-to-image translation with conditional adversarial networks.
\newblock In {\em Proceedings of the IEEE conference on computer vision and
  pattern recognition}, pages 1125--1134, 2017.

\bibitem{2014Autoencoder}
D.~P. Kingma and M. Welling.
\newblock Auto-encoding variational bayes.
\newblock {\em arXiv.org}, 2014.

\bibitem{kong2022look}
Yuxin Kong, Canjie Luo, Weihong Ma, Qiyuan Zhu, Shenggao Zhu, Nicholas Yuan,
  and Lianwen Jin.
\newblock Look closer to supervise better: One-shot font generation via
  component-based discriminator.
\newblock In {\em Proceedings of the IEEE/CVF Conference on Computer Vision and
  Pattern Recognition}, pages 13482--13491, 2022.

\bibitem{2020diffvg}
T.~M. Li, M Luká, M. Gharbi, and J. Ragan-Kelley.
\newblock Differentiable vector graphics rasterization for editing and
  learning.
\newblock {\em ACM Transactions on Graphics}, 39(6):1--15, 2020.

\bibitem{lian2018easyfont}
Zhouhui Lian, Bo Zhao, Xudong Chen, and Jianguo Xiao.
\newblock Easyfont: a style learning-based system to easily build your
  large-scale handwriting fonts.
\newblock {\em ACM Transactions on Graphics (TOG)}, 38(1):1--18, 2018.

\bibitem{implicit}
Ying-Tian Liu, Yuan-Chen Guo, Yi-Xiao Li, Chen Wang, and Song-Hai Zhang.
\newblock Learning implicit glyph shape representation.
\newblock {\em IEEE Transactions on Visualization and Computer Graphics}, 2022.

\bibitem{svgvae}
Raphael~Gontijo Lopes, David Ha, Douglas Eck, and Jonathon Shlens.
\newblock A learned representation for scalable vector graphics.
\newblock In {\em Proceedings of the IEEE/CVF International Conference on
  Computer Vision}, pages 7930--7939, 2019.

\bibitem{lyu2017auto}
Pengyuan Lyu, Xiang Bai, Cong Yao, Zhen Zhu, Tengteng Huang, and Wenyu Liu.
\newblock Auto-encoder guided gan for chinese calligraphy synthesis.
\newblock In {\em 2017 14th IAPR International Conference on Document Analysis
  and Recognition (ICDAR)}, volume~1, pages 1095--1100. IEEE, 2017.

\bibitem{mirza2014conditional}
Mehdi Mirza and Simon Osindero.
\newblock Conditional generative adversarial nets.
\newblock {\em arXiv preprint arXiv:1411.1784}, 2014.

\bibitem{park2021few}
Song Park, Sanghyuk Chun, Junbum Cha, Bado Lee, and Hyunjung Shim.
\newblock Few-shot font generation with localized style representations and
  factorization.
\newblock In {\em Proceedings of the AAAI Conference on Artificial
  Intelligence}, volume~35, pages 2393--2402, 2021.

\bibitem{reddy2021Im2Vec}
Pradyumna Reddy, Michael Gharbi, Michal Lukac, and Niloy~J Mitra.
\newblock Im2vec: Synthesizing vector graphics without vector supervision.
\newblock In {\em Proceedings of the IEEE/CVF Conference on Computer Vision and
  Pattern Recognition}, pages 7342--7351, 2021.

\bibitem{suveeranont2010example}
Rapee Suveeranont and Takeo Igarashi.
\newblock Example-based automatic font generation.
\newblock In {\em International Symposium on Smart Graphics}, pages 127--138.
  Springer, 2010.

\bibitem{tang2022fewshot}
Licheng Tang, Yiyang Cai, Jiaming Liu, Zhibin Hong, Mingming Gong, Minhu Fan,
  Junyu Han, Jingtuo Liu, Errui Ding, and Jingdong Wang.
\newblock Few-shot font generation by learning fine-grained local styles.
\newblock In {\em Proceedings of the IEEE/CVF Conference on Computer Vision and
  Pattern Recognition}, pages 7895--7904, 2022.

\bibitem{tian2017zi2zi}
Yuchen Tian.
\newblock zi2zi: Master chinese calligraphy with conditional adversarial
  networks.
\newblock {\em Internet] https://github. com/kaonashi-tyc/zi2zi}, 3, 2017.

\bibitem{2017Attentionisallyouneed}
A. Vaswani, N. Shazeer, N. Parmar, J. Uszkoreit, L. Jones, A.~N. Gomez, L.
  Kaiser, and I. Polosukhin.
\newblock Attention is all you need.
\newblock In {\em arXiv}, 2017.

\bibitem{wang2020attribute2font}
Yizhi Wang, Yue Gao, and Zhouhui Lian.
\newblock Attribute2font: Creating fonts you want from attributes.
\newblock {\em ACM Transactions on Graphics (TOG)}, 39(4):69--1, 2020.

\bibitem{wang2021deepvecfont}
Yizhi Wang and Zhouhui Lian.
\newblock Deepvecfont: synthesizing high-quality vector fonts via dual-modality
  learning.
\newblock {\em ACM Transactions on Graphics (TOG)}, 40(6):1--15, 2021.

\bibitem{xie2021dg}
Yangchen Xie, Xinyuan Chen, Li Sun, and Yue Lu.
\newblock Dg-font: Deformable generative networks for unsupervised font
  generation.
\newblock In {\em Proceedings of the IEEE/CVF Conference on Computer Vision and
  Pattern Recognition}, pages 5130--5140, 2021.

\bibitem{emd}
Yexun Zhang, Ya Zhang, and Wenbin Cai.
\newblock Separating style and content for generalized style transfer.
\newblock In {\em Proceedings of the IEEE conference on computer vision and
  pattern recognition}, pages 8447--8455, 2018.

\end{thebibliography}
}

\end{document}